\documentclass{article}
\usepackage{spconf,amsmath,graphicx}
\usepackage{amssymb}
\usepackage{booktabs}
\usepackage[dvipsnames,table]{xcolor}
\usepackage{multirow}
\usepackage{nicematrix}
\usepackage{blindtext}
\usepackage{colortbl}

\usepackage{microtype}
\newcommand{\eefgvi}{E\textsuperscript{2}FGVI}
\newcommand{\etal}{\textit{et al.}}
\usepackage{enumitem} 
\usepackage{eso-pic} 

\makeatletter
\newcommand{\btriangle}{\mathpalette\btriangle@\relax}
\newcommand{\btriangle@}[2]{%
  \begingroup
  \sbox\z@{$\m@th#1\triangle$}%
  \makebox[\wd\z@]{%
    \raisebox{0.04\height}{%
      \resizebox{1.1\wd\z@}{0.96\ht\z@}{%
        $\m@th#1\blacktriangle$%
      }%
    }%
  }%
  \endgroup
}
\makeatother
\usepackage[pagebackref,breaklinks,colorlinks,bookmarks=false]{hyperref}

\usepackage[capitalize]{cleveref}
\crefname{section}{Sec.}{Secs.}
\Crefname{section}{Section}{Sections}
\Crefname{table}{Table}{Tables}
\crefname{table}{Tab.}{Tabs.}

\title{\textsc{In2Out}: Fine-Tuning Video Inpainting Model for \\ Video Outpainting using Hierarchical Discriminator}
\name{Sangwoo Youn$^1$, Minji Lee$^2$, Nokap Tony Park$^3$, Yeonggyoo Jeon$^3$, Taeyoung Na$^3$\thanks{This research was supported by Culture, Sports, and Tourism R\&D Program through the Korea Creative Content Agency(KOCCA) grant funded by the Ministry of Culture, Sports, and Tourism(MCST) in 2024(Project Name: Development of Technology for Convergence Performance Planning and Production Platform to Revitalize the Production of Convergence Performance by Traditional Artist Dance Music, Project Number: RS-2024-00398536, Contribution Rate: 100\%)}}
\address{$^1$KAIST, $^2$Columbia University, $^3$SK Telecom}

\begin{document}
%
\maketitle
%
\AddToShipoutPictureBG*{
  \AtPageLowerLeft{
    \raisebox{20pt}{
        \hspace{1.68cm} 
        \parbox[b]{\textwidth}{
          \footnotesize 
          © 2025 IEEE. Personal use of this material is permitted. Permission from IEEE must be obtained for all other uses, in any current or future media, including reprinting/republishing this material for advertising or promotional purposes, creating new collective works, for resale or redistribution to servers or lists, or reuse of any copyrighted component of this work in other works.
        }
    }
  }
}
\begin{abstract}
Video outpainting presents a unique challenge of extending the borders while maintaining consistency with the given content. In this paper, we suggest the use of video inpainting models that excel in object flow learning and reconstruction in outpainting rather than solely generating the background as in existing methods. However, directly applying or fine-tuning inpainting models to outpainting has shown to be ineffective, often leading to blurry results. Our extensive experiments on discriminator designs reveal that a critical component missing in the outpainting fine-tuning process is a discriminator capable of effectively assessing the perceptual quality of the extended areas. To tackle this limitation, we differentiate the objectives of adversarial training into global and local goals and introduce a hierarchical discriminator that meets both objectives. Additionally, we develop a specialized outpainting loss function that leverages both local and global features of the discriminator. Fine-tuning on this adversarial loss function enhances the generator's ability to produce both visually appealing and globally coherent outpainted scenes. Our proposed method outperforms state-of-the-art methods both quantitatively and qualitatively. Supplementary materials including the demo video and the code are available in \href{https://sigport.org/documents/in2out-fine-tuning-video-inpainting-model-video-outpainting-using-hierarchical}{SigPort}.
\end{abstract}
\section{Introduction}
\label{sec:intro}



Despite the advances of diffusion models and generative adversarial networks, video outpainting has not been as extensively studied as image outpainting. Image and video outpainting are inherently distinct due to the possible existence of information about the extended region in the other frames of the video. 

In contrast, video inpainting, which involves filling in objects or free-form masks within a video in a contextually and temporally consistent manner, has been extensively studied in \cite{t-patchgan, liCvpr22vInpainting, zhou2023propainter}. Notably, recent advancements like ProPainter~\cite{zhou2023propainter} and \eefgvi\cite{liCvpr22vInpainting}, which propagate features using completed flow and reconstruct the frames through a novel architecture, have shown excellent results in both inpainting background and foreground. The ability of these models to estimate flow and reconstruct objects underscores their potential for outpainting applications. However, directly applying the inpainting model for outpainting is infeasible, producing blurry results, as pointed out in \cite{ctcvo} and our results (Fig.~\ref{fig:qual_disc_design}). While Dehan \etal~\cite{ctcvo} attribute this failure to the inherent problem of outpainting, where less surrounding information is available compared to inpainting, we, \textit{however}, attribute this to the current adversarial loss used in inpainting training. We argue that adequate fine-tuning with discriminators that assess intermediate features can successfully adapt the video inpainting model for outpainting. 

In this paper, we propose a novel approach to video outpainting, termed \textsc{In2Out}. In order to focus on the challenge of achieving both local perceptual quality and global consistency in generated video regions, we introduce a hierarchical discriminator that leverages the properties of convolutional layers.
 The early layers assess the local quality of the video, while the deeper layers evaluate global consistency by contrasting various patches of frames. To tailor the layers to the purpose, we introduce an outpainting loss function that operates on local and global features derived from real and generated videos. On the whole, our proposed adversarial framework optimizes the generator's performance in both local detail and global scene coherence. Since our idea is orthogonal to the generator architecture, our discriminator and generative loss can be used with any video inpainting model.

Our main contributions are as follows:
\begin{enumerate}[widest=0., align=left, leftmargin=*, itemsep=-3pt]
    \item An investigation, through extensive comparisons, into the failures of commonly-used discriminators in video outpainting, highlighting the critical role of the discriminator during fine-tuning;
    \item A novel adversarial objective specifically tailored for video outpainting that reduces blur in outpainted regions;
    \item The first successful adaptation of a video inpainting model to the outpainting task;
    \item Achievement of state-of-the-art performance compared to previous outpainting methods and inpainting baselines.
\end{enumerate}


\section{Related Works}


\subsection{Video Outpainting}
Video outpainting extends the contents of a video frame beyond its original boundaries while preserving the consistency of contents across neighboring frames. 
There is comparatively less work on video outpainting, and they mostly provide incomplete solutions. Lee \etal~\cite{lee2019video} warps and blends neighboring frames to extend the region based on the observed pixel, but regions that were never visible are left blank. While some video inpainting methods \cite{gao2020flow} evaluate their methods in video outpainting as well, they perform worse than inpainting.

\textbf{Background-only methods}. Dehan \etal~\cite{ctcvo} use a video object segmentation (VOS) network to detach objects from the background and then employ a flow-based video completion network to generate the background. 
Jin \etal~\cite{wacvw} similarly employ a VOS network but stretch the background rather than generating content.

\textbf{Generative methods}. Fan \etal~\cite{m3ddm} propose a masked 3D diffusion model with classifier-free guidance \cite{ho2022classifierfree} to tackle video outpainting.
Recently, Wang \etal~\cite{wang2025your} introduce a diffusion based pipeline comprises input-specific adaptation and pattern-aware outpainting for video outpainting.
However, diffusion based methods\cite{m3ddm,wang2025your} are limited to process only particular sizes of videos.

\subsection{Discriminators in Image/Video Inpainting}

Discriminator and generative loss are widely used in image and video inpainting to enhance the perceptual quality of the generated results. Pathak \etal~\cite{pathak2016context} first propose to use adversarial loss to alleviate blurry results caused by the pixel-wise reconstruction loss in image inpainting. They use \textit{global} discriminator that looks at an entire image to evaluate the consistency between generated features and real features. To further focus on the perceptual quality of the generated region, Iizuka \etal~\cite{iizuka2017globally} propose to use \textit{partial} discriminator that looks only at the inpainted region together with the global discriminator. Due to the inapplicability of partial discriminator in free-form inpainting, where mask can exist anywhere in any shape, Yu \etal~\cite{yu2019free} propose to apply adversarial loss on the feature maps of the discriminator, instead of the single predicted log likelihood value. Chang \etal~\cite{t-patchgan} extends this \textit{global feature} loss to temporal dimension, by using 3-dimensional convolution. This T-PatchGAN discriminator and loss is widely used in video inpainting \cite{zeng2020learning, liCvpr22vInpainting, liu2021fuseformer,zhang2022flow,zhou2023propainter}.

\section{Proposed Method}
\label{sec:method}

In this section, we propose \textit{hierarchical discriminator} driven from the failures of existing discriminators, and formulate the video outpainting loss.

The spatio-temporal feature discriminator $\mathcal{D}$ learns to classify each patch of given video as real or fake. Given real video, $x\sim P_Y (x)$ and video generated by generator $\mathcal{G}$, $z\sim P_{\hat{Y}}(z)$, the general training objective of discriminator is a hinge loss on model output,
\begin{align*}
    \mathcal{L}_\mathcal{D} &= \mathbb{E}_{x\sim P_Y (x)} [\text{ReLU} (1-\mathcal{D}(x))]  \\ 
    &+ \mathbb{E}_{z\sim P_{\hat{Y}} (z)} [\text{ReLU} (1+\mathcal{D}(z))].
\end{align*}

\noindent
This objective aims to maximize the margin between the real patches and fake patches. The inpainting generator $\mathcal{G}$ is typically trained on multiple objectives including reconstruction loss $\mathcal{L}_\text{rec}$ and adversarial loss $\mathcal{L}_\text{adv}$,
\begin{align*} \label{gen-loss}
\mathcal{L}_\mathcal{G} &= \lambda_\text{rec} \mathcal{L}_\text{rec} + \lambda_\text{adv} \mathcal{L}_\text{adv} + \lambda_\text{flow} \mathcal{L}_\text{flow},\\
\mathcal{L}_\text{rec} &=||\hat{Y}-Y ||_1,\\
\mathcal{L}_\text{adv} &= -\mathbb{E}_{z \sim P_{\hat{Y}}(z)}[\mathcal{D}(z)].
\end{align*}

Our work focuses on training the discriminator $\mathcal{D}$ and defining the loss functions $\mathcal{L}_\mathcal{D}$ and $\mathcal{L}_\text{adv}$ in a way that effectively adapts the \textit{inpainting} generator $\mathcal{G}_{\theta_\text{in}}$ to the \textit{outpainting} generator $\mathcal{G}_{\theta_\text{out}}$.

\subsection{Hierarchical Discriminator}

The purpose of the discriminator can be divided into two: (i) ensuring \textbf{global} consistency of the scene, and (ii) ensuring \textbf{local} perceptual quality of generated region. The T-PatchGAN discriminator primarily targets the former, a global objective, as per the design in which the receptive field of the last convolutional layer covers an entire video. However, the discriminator that effectively evaluates the local quality of the outpainted region is necessary, especially in an outpainting setting. 

While the approach proposed by Iizuka \etal~\cite{iizuka2017globally}, which involves the training of two distinct discriminators for each of these objectives, might seem viable, it is fraught with its own set of challenges. Managing multiple GAN losses introduces a delicate balance and sensitivity to hyperparameters, often resulting in training failures (See Sec.~\ref{sec:quant_discs}). 

\begin{figure}[t!]
    \centering
     \includegraphics[width=\columnwidth]{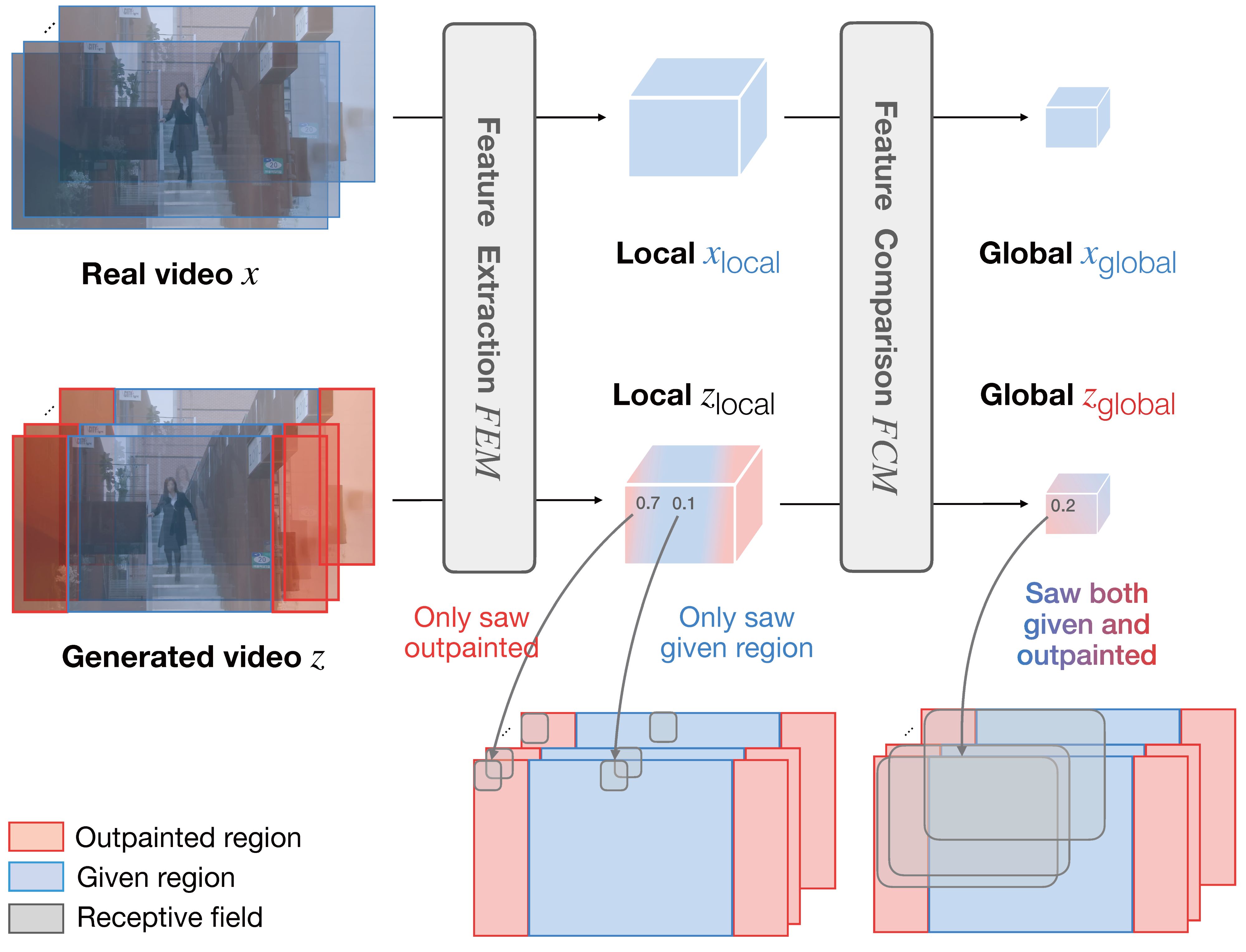}
     \caption{\textbf{Outpainting loss calculation using hierarchical discriminator}. We employ the local and global features, which are the output of the feature extraction module (FEM) and feature comparison module (FCM), respectively. $\texttt{Out}(z_\text{local})$ in Eq.~\ref{eq:outpainting-loss} only saw outpainted region (\textcolor{red}{red}).}
     \label{fig:disc_arch}
\end{figure}

We believe that a single discriminator can effectively capture both local and global objectives in outpainting scenarios by cleverly leveraging the properties of convolutional layers. As layers progress deeper, the receptive field of features expands, enabling the local features to have a smaller receptive field while the global features perceive the entire video. Thus, the receptive field of earlier layers is restricted to the outpainted region for the pixels at the side, and to the generated region for the pixels in the center, as illustrated in Fig.~\ref{fig:disc_arch}. Motivated by this, we propose a \textit{hierarchical discriminator} $\mathcal{D}_\text{hierarchical}$ where the initial layers, termed feature extraction module (FEM), focus on assessing the local quality of video, whereas deeper layers, termed feature comparison module (FCM), focus on comparing the different patches of frames and assess global consistency. (See Fig.~\ref{fig:disc_arch}) 

\subsection{Outpainting Loss}

We enforce these unique roles of modules via our proposed \textit{outpainting loss} (See Eq.~\ref{eq:outpainting-loss}) that operates on both local features $x_\text{local}, z_\text{local}$ and global features $x_\text{global}, z_\text{global}$. Given real video, $x\sim P_Y (x)$ and video generated by generator $\mathcal{G}$, $z\sim P_{\hat{Y}}(z)$, the hierarchical discriminator compute features, 

\begin{align*}
x_\text{local} &= \text{FEM} (x), z_\text{local} = \text{FEM}(z), \\
x_\text{global} &= \text{FCM} (x_\text{local}), z_\text{global} = \text{FCM} (z_\text{local}).
\end{align*}

The outpainting loss is defined as 
\begin{equation} \label{eq:outpainting-loss}
\begin{split}
\mathcal{L}_\text{out} = &\mathop{\mathbb{E}_{x \sim P_{Y}(x)}}[\alpha_\text{local} \cdot \text{ReLU}(1-x_\text{local}) \\ &+ \alpha_\text{global} \cdot \text{ReLU}(1-x_\text{global})] + \\
&\mathop{\mathbb{E}_{z \sim P_{\hat{Y}}(z)}}[\alpha_\text{local} \cdot \text{ReLU}(1+\texttt{Out}(z_\text{local})) \\ &+ \alpha_\text{global} \cdot \text{ReLU}(1+z_\text{global})].
\end{split}
\end{equation}

Let mask ratio $m$. For simplicity, we define procedure \texttt{Out} which indicates the outpainted region of video, \emph{i.e.} $\texttt{Out}(x) = x({i,j})$ such that $i<(m/2)\cdot \texttt{width}(x)$ or $i>(1-m/2)\cdot \texttt{width}(x)$. Here, note that we use $\texttt{Out}(z_\text{local})$ instead of $z_\text{local}$. Since the FEM has a small receptive field, $z_\text{local}$ contains local information of generated inputs. Thus, the discriminator should not be trained to classify the center of $z_\text{local}$, which is the feature of given region, as fake. The mapping \texttt{Out} only reflects calculations for the outpainted regions for the local feature. Specifically, FEM is designed to have a receptive field size identical to the size of the outpainted region.

For the video inpainting generator, the adversarial loss is defined as:
\begin{equation} \label{loss}
\begin{split}
\mathcal{L}_\text{adv} = &\mathop{\mathbb{E}_{z \sim P_{\hat{Y}}(z)}}[\text{FCM}(\text{FEM}(z))].
\end{split}
\end{equation}

\begin{table*}[t!]
    \centering
    \setlength{\tabcolsep}{5pt}
    \scalebox{0.9}{
    \begin{NiceTabular}{ll|cll|cll}
    \toprule
         & \multirow{2}{*}{Model} & \multicolumn{3}{c}{Youtube-VOS} & \multicolumn{3}{c}{DAVIS}\\
        & & PSNR $\uparrow$ & SSIM $\uparrow$ & VFID $\downarrow$ & PSNR $\uparrow $ & SSIM $\uparrow$ & VFID $\downarrow$ \\
         \toprule
         
        \textsc{Background-only} & Dehan \etal~\cite{ctcvo} & 21.99 & 0.8632 & 0.085 & 25.78 & 0.8901 & \textbf{0.104} \\
         \midrule
        \multirow{2}{*}{\textsc{Diffusion}}  
        & M3DDM\cite{m3ddm} & 24.16 & 0.8862 & 0.091 & 24.64 & 0.8641 & 0.187 \\ 
        & MOTIA\cite{wang2025your} & 22.95 & 0.8795 & 0.208 & 24.51 & 0.8624 & 0.177 \\ 
        \midrule
        \multirow{3}{*}{\textsc{Inpainting}} & FuseFormer \cite{liu2021fuseformer} & 23.78 & 0.7899 & 0.098 & 25.55 & 0.7861 & 0.193 \\
        & ProPainter\cite{zhou2023propainter} & 22.74 & 0.9292 & 0.097 & 25.14 & 0.9353 & 0.144\\
        & \eefgvi \cite{liCvpr22vInpainting} & 23.81 & 0.9378 & 0.093 & 24.73 & 0.9290 & 0.158 \\
          \midrule
         \multirow{2}{*}{\textbf{\textsc{In2Out} (Ours)}} 
         & ProPainter &  25.18 \small{(\textcolor{red}{$\scriptstyle\btriangle$2.4})} & 0.9399 & \textbf{0.075} & \textbf{27.33} \small{(\textcolor{red}{$\scriptstyle\btriangle$2.2})} & \textbf{0.9431} & 0.115 \\
         & \eefgvi & \textbf{25.71} \small{(\textcolor{red}{$\scriptstyle\btriangle$1.9})} & \textbf{0.9464} & 0.096 & 26.61 \small{(\textcolor{red}{$\scriptstyle\btriangle$1.9})} & 0.9385 & 0.139 \\
         \bottomrule
    \end{NiceTabular}}
    \caption{\textbf{Quantitative comparisons} on Youtube-VOS and DAVIS datasets. Mask ratio is set to $1/4$. $\uparrow$ indicates higher is better, and $\downarrow$ indicates lower is better. The \textcolor{red}{value} in the parentheses indicate the increase in PSNR by fine-tuning inpainting models using our discriminator.}
    \label{tab:quant_main}
\end{table*}

\section{Results}
\label{expres}
In this section, we compare our method with several state-of-the-art video outpainting methods and demonstrate the impact of discriminator design on outpainting performance.

\subsection{Settings}
\label{sec:exp_settings}

\textbf{Models.} Our method is agnostic to the specific generator used. To demonstrate its effectiveness, we fine-tune two video inpainting models with our discriminator and loss: ProPainter~\cite{zhou2023propainter} and \eefgvi~\cite{liCvpr22vInpainting}, and report the performance improvements. 
The training details are included in the supplementary material.
We compare our method with the background-only~\cite{ctcvo} and diffusion-based~\cite{m3ddm,wang2025your} video outpainting models. Additionally, we present the baseline performance of the video inpainting model FuseFormer~\cite{liu2021fuseformer}. Note that FuseFormer, M3DDM, and MOTIA can only process videos with a resolution of 240p, 144p, and 256p, respectively. Therefore, we downsampled the input videos and upsampled the generated videos when evaluating their performance.

\textbf{Datasets.}
To assess the performance of our proposed approach, we conduct evaluations on two recognized video datasets: YouTube-VOS \cite{vos2018} and DAVIS \cite{Perazzi2016}. 
For DAVIS, following Liu \etal~\cite{liu2021fuseformer}, we evaluate in 50 video clips from the test set. The videos of DAVIS dataset we evaluated are 480p. We fine-tuned our model on the train set of YouTube-VOS dataset resized to 240p. During the evaluation, we used the test set of Youtube-VOS dataset resized to 360p.




\textbf{Metrics.} We choose Peak Signal To Noise Ratio (PSNR), structural similarity index measure (SSIM) \cite{ssim}, and Video Frchet Inception Distance (VFID) \cite{vfid} to evaluate the quality of the outpainted videos. Note that we compute the metrics in whole video, not only the outpainted region. VFID measures the perceptual similarity between two input videos using a pretrained I3D \cite{i3d} model and has been widely used in recent video inpainting works.


\subsection{Quantitative Results}
\label{sec:main_res}

As shown in Tab.~\ref{tab:quant_main}, our method demonstrates superior reconstruction performance on both Youtube-VOS and DAVIS datasets compared to SOTA models. Notably, on Youtube-VOS dataset, \eefgvi\cite{liCvpr22vInpainting} adapted to outpainting using our \textsc{In2Out} method attains a PSNR 1.9dB higher than the baseline and 3.7dB higher than Dehan \etal. 
We also outperform M3DDM \cite{m3ddm} and MOTIA \cite{wang2025your} with a large margin. 
These results demonstrate that our method excels in outpainting and successfully adapts the inpainting model for outpainting. 

\begin{figure}[t!]
    \centering
    \includegraphics[width=\columnwidth]{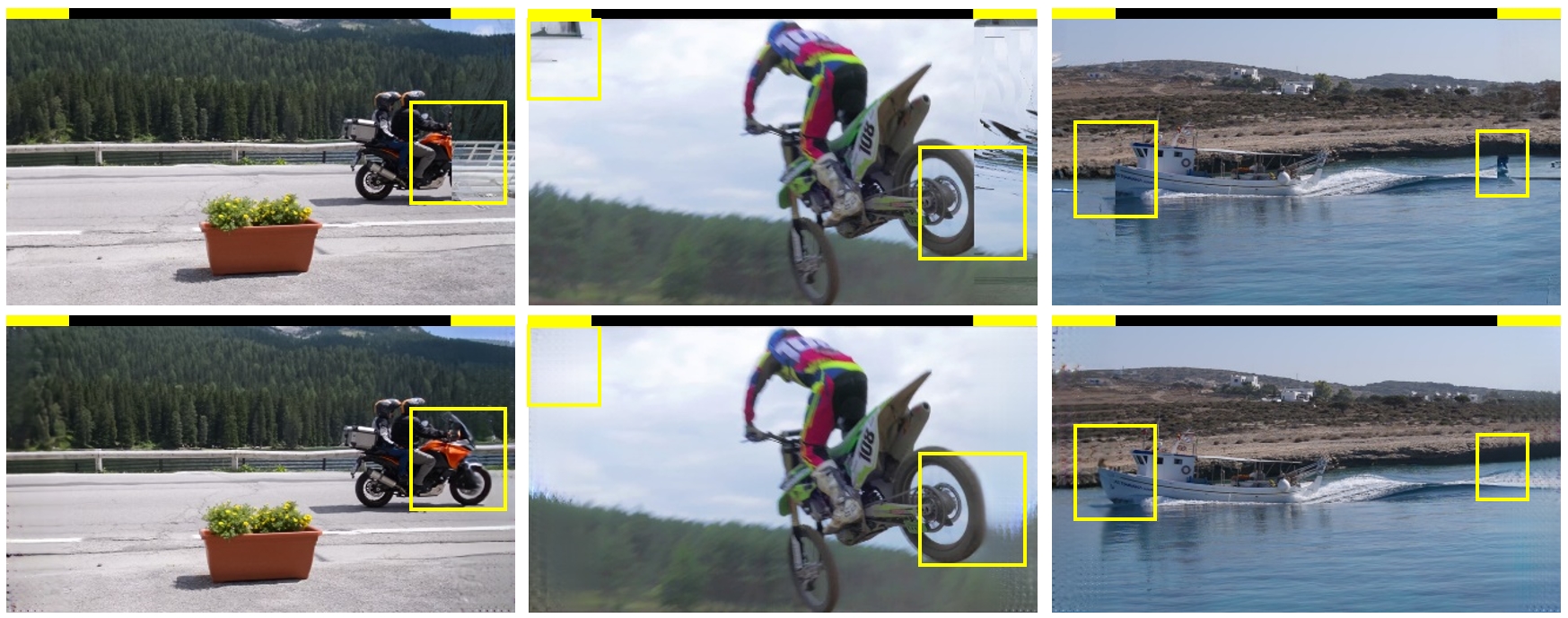}

    \caption{\textbf{Qualitative comparisons of Dehan \etal~\cite{ctcvo} (top) and our \textsc{In2Out} fine-tuned \eefgvi~(bottom)} on 480p DAVIS dataset.}
    \label{fig:qual_ctcvo}
\end{figure}

\begin{figure}[t!]
    \centering
    \includegraphics[width=\columnwidth]{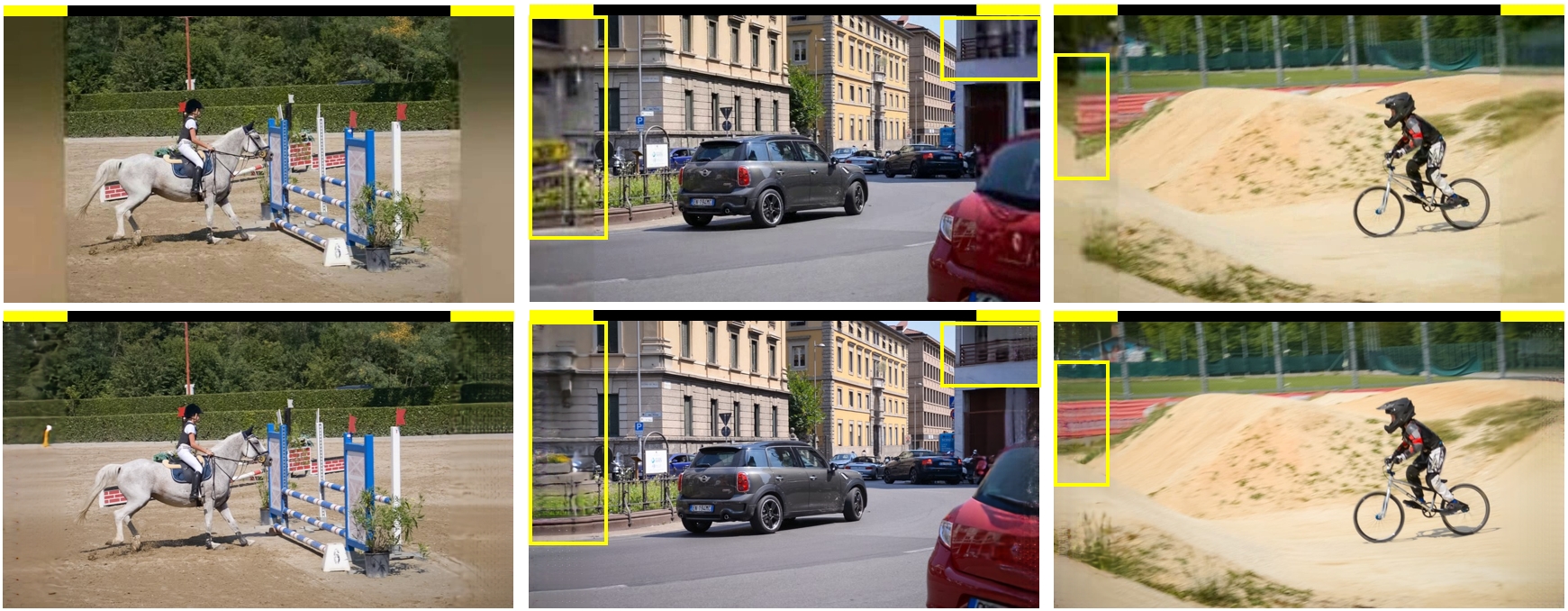}

    \caption{\textbf{Qualitative comparisons of M3DDM \cite{m3ddm} (top) and our \textsc{In2Out} fine-tuned \eefgvi~(bottom)} on 480p DAVIS dataset.}
    \label{fig:qual_m3ddm}
\end{figure}

\subsection{Qualitative Results}
\label{sec:res_qual}

The yellow line on the top of the video indicates the horizontally outpainted region. Figure~\ref{fig:qual_ctcvo} visually compares the three outpainted videos of Dehan~\etal~\cite{ctcvo} and our method. In the outpainted video by Dehan~\etal, objects are truncated, whereas our method seamlessly completes objects moving into the outpainted area. 

M3DDM fails to generate content and produces blurry results in 15 out of 50 videos from the DAVIS test dataset. Examples of these failures are shown in the leftmost image in Fig.~\ref{fig:qual_m3ddm}. Even in videos where M3DDM successfully outpaints, our method provides more accurate and complete results, as shown in the two right images in Fig.~\ref{fig:qual_m3ddm}.

As shown in Fig.~\ref{fig:qual_disc_design}, the baseline inpainting model produces blurry results, especially at the boundaries of the frames. Our method allows the generator to adapt to outpainting, and significantly reduce blurry artifacts. 

\subsection{Efficiency}

\label{sec:res_efficiency}

\begin{table*}[t!]
    \centering
    \scalebox{0.9}{
    \begin{NiceTabular}{l|ccc|ccc}
    \toprule
        & \multicolumn{3}{c}{Youtube-VOS} & \multicolumn{3}{c}{DAVIS}\\
         Discriminator & PSNR $\uparrow$ & SSIM $\uparrow$ & VFID $\downarrow$ & PSNR $\uparrow $ & SSIM $\uparrow$ & VFID $\downarrow$ \\
         \midrule
         None & 24.53 & 0.9256 & 0.115 & 24.51 &0.8984 & 0.220 \\
         Global (T-PatchGAN \cite{t-patchgan})  &24.28 & 0.9237 &0.107 & 24.04 & 0.8958 & 0.166\\
         Partial-only & 24.04 & 0.9083 & 0.086 & 25.67 & 0.9272 & 0.164 \\
         Global \& partial \cite{iizuka2017globally} & 13.11 & 0.7869  & 0.181 & 12.79 & 0.7709 & 0.317 \\
         Local-only &24.47 & 0.9179 &\textbf{0.082} &25.74 &0.9322 & 0.162 \\
         \textbf{Hierarchical (Ours)} & \textbf{25.71} & \textbf{0.9464} & 0.096 & \textbf{26.61} & \textbf{0.9385} & \textbf{0.139} \\
         \bottomrule
    \end{NiceTabular}}
    \caption{\textbf{Quantitative comparisons of discriminator designs} on Youtube-VOS and DAVIS datasets. Mask ratio is set to $1/4$. $\uparrow$ indicates higher is better, and $\downarrow$ indicates lower is better.}
    \label{tab:quant_by_disc}
\end{table*}

Dehan \etal takes about 21s/frame to outpaint a 480p video, due to their iterative ouptainting scheme. In contrast, our approach employing the end-to-end inpainting model takes about 0.4s/frame over 52 times faster than theirs.  
Additionally, M3DDM and MOTIA takes about 5s/frame and 24.3s/frame, respectively, even though M3DDM operates on 144p video and MOTIA operates on 256p video, which is order of magnitude slower than inpainting models. 
This underscores that employing a video inpainting approach for outpainting is a promising and effective strategy.

\subsection{Discriminator Designs}

In this section, we extensively study the effect of discriminator design on the outpainting adaption of video inpainting model. We used \eefgvi~for the inpainting generator. Table~\ref{tab:quant_by_disc} compares the designs of each discriminator, starting with \emph{None} where no discriminator is employed during fine-tuning. Other designs include: \emph{Global} where the discriminator processes the entire video and utilizes only the final output to calculate the loss (equivalent to T-PatchGAN discriminator); \emph{Partial-only} that exclusively processes the outpainted region and considers only the final output of the discriminator to evaluate the loss; \emph{Global \& partial} that averages the losses from both discriminators to compute the total discriminator loss (equivalent to the discriminator proposed by Iizuka \etal~\cite{iizuka2017globally})); and lastly, \emph{Local-only} that takes the full video as input but exclusively utilizes the local features $x_\text{local}$ and \texttt{Out}($z_\text{local}$) of the discriminator to determine the loss.

\subsubsection{Quantitative Results}

As shown in Tab.~\ref{tab:quant_by_disc}, all other discriminator designs show a decrease in performance compared to fine-tuning without discriminator, except our proposed design, on Youtube-VOS dataset. The trend is similar on DAVIS dataset, but partial-only and local-only discriminators outperform the fine-tuning without discriminator. The success of discriminators focusing on the quality of outpainted regions highlights the importance of assessing local quality in the outpainting task. Global \& partial discriminator shows the most severe performance degradation. We observed from the log that the loss fails to converge, and thus the training fails, underscoring the infeasibility of balancing multiple discriminators targeting different objectives. 

\subsubsection{Qualitative Results}
\label{sec:quant_discs}
\begin{figure}[t!]
    \centering
    \includegraphics[width=0.97\columnwidth]{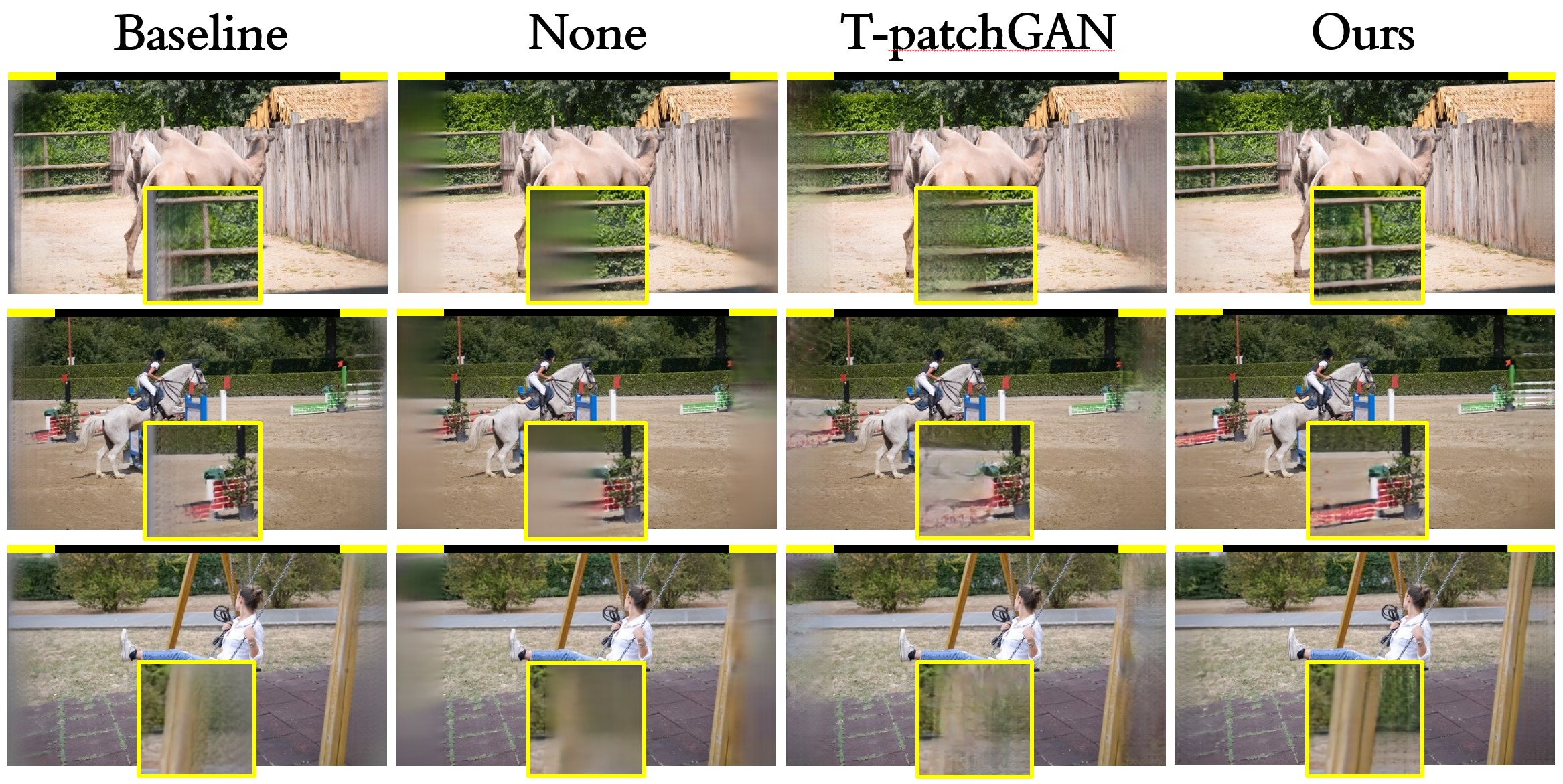}
    \caption{\textbf{Qualitative comparisons of discriminator designs} on 480p DAVIS dataset.}
    \label{fig:qual_disc_design}
\end{figure}

We also show the visual comparison of the results of the different discriminator designs in Fig.~\ref{fig:qual_disc_design}. The fine-tuning without a discriminator and the fine-tuning with the global discriminator led to more blurry results compared to the baseline. 
Compared to other designs, our hierarchical discriminator achieves the most accurate and consistent restoration of foreground and least blurry artifacts, demonstrating the effectiveness of our proposed outpainting loss that considers both local and global objectives in increasing the perceptual quality of the outpainted region.

\subsection{Ablation Study on Mask Ratio}
\label{sec:ablation}


\begin{table}[h]
    \centering
    \scalebox{0.9}{
    \begin{NiceTabular}{lcc}
    \toprule
         & \multicolumn{3}{c}{Ratio} \\
         Method & $1/3$ & $1/6$ \\
    \midrule
         Dehan \etal~\cite{ctcvo} & 23.34 / 0.8234 & 29.46 / 0.9359\\
         M3DDM \cite{m3ddm} & 21.92 / 0.8146 & 27.53 / 0.9133\\
         \eefgvi \cite{liCvpr22vInpainting} & 22.13 / 0.8790 & 28.54 / 0.9385 \\
         ProPainter & 22.61/ 0.8792 & 28.30 / 0.9736\\
         \textbf{Ours} (\eefgvi) & 23.94 / \textbf{0.9251} & 30.05 / 0.9765\\
         \textbf{Ours} (ProPainter) & \textbf{24.72} / 0.8916 & \textbf{30.70} / \textbf{0.9780}\\
    \bottomrule
    \end{NiceTabular}}
    \caption{\textbf{Comparison of PSNR/SSIM by mask ratios} on DAVIS dataset. See Supplementary Sec.~\ref{sec:suppl_extended} for the full comparison including VFID metrics.}
    \label{tab:ablation_ratio}
\end{table}

Table~\ref{tab:ablation_ratio} presents the outpainting performance at varying mask ratios. As the mask ratio increases, the task becomes more challenging, resulting in generally lower performance at a ratio of $1/3$ and improved performance at $1/6$. In both scenarios, our method achieves higher PSNR and SSIM compared to Dehan \etal, M3DDM, \eefgvi~baseline and ProPainter baseline. This consistently high performance highlights the robustness of our fine-tuning approach.

\section{Conclusion}
In this work, we propose a novel adversarial framework to fine-tune video inpainting generators to outpainting, paving the effective way to exploit the powerful priors for a task that is comparatively less studied. The contribution also lies in that we extensively compare different discriminator designs, and suggest that the discriminator enforcing both local and global objective may be the missing piece of the successful adaptation of inpainting to outpainting. Our experiments demonstrate that our proposed method outperforms existing video outpainting models in terms of quantitative and qualitative measures. Notably, our discriminator can be integrated into any existing video inpainting model, providing a solid starting point for future research in this domain.


\bibliographystyle{IEEEbib}
\bibliography{egbib}

\clearpage

\setcounter{section}{0}
\setcounter{page}{1}

\section{Training Details}

 We used \eefgvi~HQ \cite{liCvpr22vInpainting} and ProPainter \cite{zhou2023propainter} for a baseline pre-trained generator. 
The generator and discriminator are trained simultaneously using Adam optimizer for $5 \cdot 10^4$ iterations. Learning rate is set to $4 \cdot 10^{-5}$ for both models. For \eefgvi, we set $\lambda_\text{rec}=\lambda_\text{valid}=1, \lambda_\text{flow}=0.01, \lambda_\text{adv}=0.04$, and $\alpha_\text{local}=\alpha_\text{global}=0.5$. For ProPainter, we set the values same as \eefgvi~except $\lambda_\text{flow}=1$ and  $\lambda_\text{adv}=0.01$.
During training, all frames are resized into 432 $\times$ 240 and the number of local frames and non-local frames (See \eefgvi~\cite{liCvpr22vInpainting}) are set to 5 and 3, respectively. Training took approximately 390 hours on one RTX 4090 GPU when fine-tuning \eefgvi. During evaluation and test, following the previous practices, we use sliding window with the size of 10.

\textbf{Masks.} 
While our primary target is outpainting 4:3 videos to 16:9 videos ($m=1/4$), we fine-tuned the generator to mask ratio of minimum $1/12$ to maximum $1/3$ to increase robustness of the model.

\textbf{Model architecture.}
For FEM, we stack three 3D convolutional layers with a spatial stride size of 2. The receptive field is $\approx 2^3 \cdot 7 = 56$ which is similar to the width of the outpainted region when mask ratio $m=1/4$, 54. For FCM, we also stack three 3D convolutional layers with a spatial stride size of 2. The receptive field is $\approx 2^6 \cdot 7 = 448$ which is larger than the width of the training data, 432. 

\section{Extended Results}
\label{sec:suppl_extended}
Here we present the VFID results of Tab.~\ref{tab:ablation_ratio}.
\begin{table}[h!]
    \centering
    \begin{NiceTabular}{l|c|c}
    \toprule
         Method & $1/3$ & $1/6$ \\
    \midrule
        Dehan \etal~\cite{ctcvo} & 0.130 & 0.071\\
        M3DDM \cite{m3ddm} & 0.277 & 0.120\\
        \eefgvi \cite{liCvpr22vInpainting} & 0.217 & 0.095 \\
        ProPainter\cite{zhou2023propainter} & 0.193 & 0.105 \\
        Ours (\eefgvi) & 0.204 & 0.092\\
        Ours (ProPainter) & 0.156 & 0.075\\
    \bottomrule
    \end{NiceTabular}
    \caption{\textbf{VFID by the outpainting ratios} on the DAVIS dataset.}
    \label{tab:full_ablation_ratio}
\end{table}

\section{Extended Ablation Studies}
\subsection{Ablation on Additional Generator}
\label{supp:any_gen}

\begin{table}[h!]
    \centering
    \begin{NiceTabular}{lcccc}
    \CodeBefore
      \rowcolor{gray!15}{5}
    \Body
    \toprule
        Discriminator  & PSNR & SSIM & VFID \\
    \midrule
        w/o Fine-tuning & 25.55 & 0.7861 & 0.193 \\
        T-PatchGAN \cite{t-patchgan} & 26.06 & 0.7907 & \textbf{0.167}\\
        Ours & \textbf{26.24} & \textbf{0.7916} & 0.177 \\
    \bottomrule
 \end{NiceTabular}
    \caption{\textbf{Quantitative comparison of discriminator design} on DAVIS dataset and FuseFormer \cite{liu2021fuseformer} generator.}
    \label{tab:disc_other_gens}
\end{table}

As shown in Tab.~\ref{tab:disc_other_gens}, our fine-tuning framework increases the performance of FuseFormer \cite{liu2021fuseformer} in both PSNR and SSIM metrics, compared to the T-PatchGAN discriminator. Thus, effectiveness of our method is not restricted to \eefgvi\cite{liCvpr22vInpainting} and ProPainter\cite{zhou2023propainter}, and can be used with any video inpainting model.  

\subsection{Flow loss weight}
\begin{table}[h!]
    \centering
    \begin{NiceTabular}{cc|ccc}
    \CodeBefore
      \rowcolor{gray!15}{2}
    \Body
    \toprule
         $\lambda_\text{gen}$ & $\lambda_\text{flow}$& PSNR $\uparrow $ & SSIM $\uparrow$ & VFID $\downarrow$ \\
    \midrule
        1 & 0.01 & 26.61 & 0.9385 & 0.139 \\
        1 & 0.1 & 26.43 & 0.9375 & 0.146 \\
        1 & 1.0 & 26.26 & 0.9363 & 0.147 \\
    \bottomrule
    \end{NiceTabular}
    \caption{\textbf{Ablation study on the flow loss weight} on the DAVIS dataset. Note that \eefgvi~baseline is trained to $\lambda_\text{flow}=1$.}
    \label{tab:flow_weight}
\end{table}

As shown in Tab.~\ref{tab:flow_weight}, lower flow weight in generator loss led to a slight increase in all metrics. This is expected since the inpainting task that incorporates object mask during training is better for learning the flow estimation. 

\subsection{Generative loss weight}

\begin{table}[h!]
    \centering
    \begin{NiceTabular}{cc|ccc}
    \CodeBefore
      \rowcolor{gray!15}{4}
    \Body
    \toprule
         $\alpha_\text{inter}$ & $\alpha_\text{global}$ & PSNR $\uparrow $ & SSIM $\uparrow$ & VFID $\downarrow$ \\
    \midrule
        0.9 & 0.1 & 26.50 & 0.9383 & 0.149 \\
        0.1 & 0.9 & 26.31 & 0.9365 & 0.137  \\
        0.5 & 0.5 & 26.61 & 0.9385 & 0.139 \\
    \bottomrule
    \end{NiceTabular}
    \caption{\textbf{Ablation study on the local and global loss weight} on the DAVIS dataset.}
    \label{tab:full_flow_weight}
\end{table}

As shown in Tab.\ref{tab:full_flow_weight}, different configurations of hyperparameters do not markedly affect the performance in all metrics, highlighting the robustness of our method to hyperparameters.

\newpage
\section{Limitation}

\begin{figure}[!h]
    \centering
    \includegraphics[width=\linewidth]{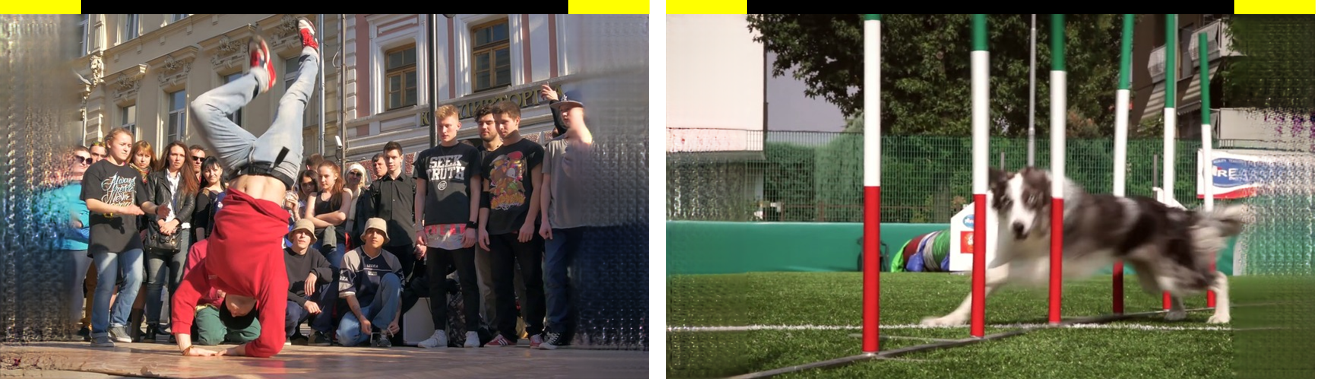}
    \caption{\textbf{Failure cases when outpainting static videos} in 480p DAVIS dataset. The yellow line on the top of the video indicates the horizontally outpainted region.}
    \label{fig:limitation}
\end{figure}

Fig.~\ref{fig:limitation} shows the failure case when outpainting static video. Our method sometimes blurs (left) or omits (right) the foreground that is never seen in a given region. This shows the continuing challenge of static videos in video outpainting.

\end{document}